\documentclass[a4paper, 10pt, conference]{ieeeconf}      

\IEEEoverridecommandlockouts                              

\usepackage{graphicx} 
\usepackage{epsfig} 
\usepackage{mathptmx} 
\usepackage{times} 
\usepackage{amsmath} 
\usepackage{amssymb}  
\usepackage{mathtools}
\usepackage{xcolor}
\usepackage{gensymb}
\usepackage{url}
\usepackage{subfigure}
\usepackage{gensymb}

\usepackage[firstpage]{draftwatermark}
\SetWatermarkText{This paper has been accepted for publication at IEEE 22nd Intelligent Transportation Systems Conference (ITSC), Auckland, New Zealand, 2019. @IEEE}
\SetWatermarkVerCenter{3cm}
\SetWatermarkAngle{0}
\SetWatermarkFontSize{8pt}
\SetWatermarkColor[gray]{0.4}
\SetWatermarkScale{0.9}

\title{\LARGE \bf  Accurate IMU Preintegration Using Switched Linear Systems For Autonomous Systems}

\author{John Henawy$^{1,*}$, Zhengguo Li$^{2,*}$,  Wei Yun Yau$^2$ , Gerald Seet$^3$ and Kong Wah Wan$^{2}$
\thanks{$^*$Joint first authors.}
\thanks{$^{1}$John Henawy is with the School of Mechanical and Aerospace Engineering,
       Nanyang Technological University, Singapore, 639798 and Institute for Infocomm Research, A-STAR, Singapore, 138632 and is a recipient of the A*STAR SINGA scholarship.
        {\tt\small johnfari001@e.ntu.edu.sg}}%
\thanks{$^{2}$Zhengguo Li , Wei Yun Yau and Kong Wah Wan are with the Institute for Infocomm Research, A-STAR, Singapore, 138632
        {\tt\small \{ezgli,wyyau,kongwah\}@i2r.a-star.edu.sg}}%
\thanks{$^{3}$Gerald Seet is with the School of Mechanical and Aerospace Engineering,
       Nanyang Technological University, Singapore, 639798
       {\tt\small MGLSEET@ntu.edu.sg}}%
}

\begin{document}


\maketitle
\thispagestyle{empty}
\pagestyle{empty}

\begin{abstract}

Employing an inertial measurement unit (IMU) as an additional sensor can dramatically improve both reliability and accuracy of visual/Lidar odometry (VO/LO). Different IMU integration models are introduced using different assumptions on the linear acceleration from the IMU. In this paper, a novel  IMU integration model is proposed by using switched linear systems. The proposed approach assumes that both the linear acceleration and the angular velocity in the body frame are constant between two consecutive IMU measurements. This is more realistic in real world situation compared to existing approaches which assume that linear acceleration is constant in the world frame while angular velocity is constant in the body frame between two successive IMU measurements. Experimental results show that the proposed approach outperforms the state-of-the-art IMU integration model. The proposed model is thus important for localization of high speed autonomous vehicles in GPS denied environments.

\end{abstract}

\section{INTRODUCTION}

State estimation is definitely one of the most essential modules in many applications such as an autonomous driving, self-localization systems and precise navigation \cite{javanmardi2017autonomous}, \cite{murali2017utilizing}. Environment which has poor or lack of global positioning system (GPS) represents a great challenge for motion estimation. Accordingly, motion estimation of an agent such as an unmanned aerial vehicle (UAV) or an unmanned ground vehicle (UGV) is crucial in such environment \cite{valenti2018enabling}, \cite{trawny2007vision}. Employing an inertial measurement unit (IMU) as an additional sensor could allow low latency high update rate state output required for low-level control \cite{li2019asymptotic} and simplify visual odometry (VO) \cite{jones2011visual}, \cite{abdi2016pose} and lidar odometry (LO) \cite{hemann2016long}, \cite{zhang2017low}, \cite{kumar2017lidar}. The estimated position, velocity and orientation can be obtained by integrating both the linear acceleration and angular velocity from the IMU assuming that the discrete outputs are constant between two successive sampling points. However, the accumulated error due to the integration process induces substantial drift and inaccuracy in attitude estimation. Sensor fusion can be adopted to correct the drift whereby the estimated state obtained by the motion integration using the IMU data is used as a prediction state to update or correct the estimated state using another onboard sensor \cite{5427255}, \cite{perez2016fusion}.

Sensor fusion techniques are undoubtedly crucial for online state estimation. Graph based optimization approaches give higher state estimation accuracy than the filtering based approaches \cite{grisetti2010tutorial}, \cite{shen2015tightly}, \cite{yang2017monocular}, \cite{mourikis2007multi} and \cite{brossard2018invariant}. However, the fact that the IMU has a higher sampling rate than other sensors is used in existing motion estimation represents a great challenge for online estimation using graph based optimization approaches. Whereby, the trajectory grows rapidly and the online estimation becomes infeasible. Lupton and Sukkarieh \cite{6092505} established the pre-integration theory whereby the linear acceleration and angular velocity are integrated in the body frame before they are transferred to the world frame between two consecutive keyframes. However, previously both the linear acceleration and the angular velocity were transferred to the world frame at each time stamp to estimate the state before being integrated. Forster et al. \cite{7557075} has introduced the  IMU preintegration on the manifold SO(3) on top of the work in  \cite{6092505}. The IMU preintegration theory is widely used nowadays because it integrates the IMU mesaurements into a single value between two consecutive frames. Thus, the online optimization can be done because the graph-based optimization have been simplified. However, the IMU integration model has been obtained under different assumptions to solve the differential equations whereby {\it the linear acceleration in the world frame and the angular velocity in the body frame are assumed to be constant between two consecutive IMU measurements in \cite{7557075}}. Consequently, the agent orientation has also been assumed to be constant for each integration interval between two successive IMU measurements. However, this assumption is not always true. For example, the linear acceleration of a multi-copter system has a higher chance to be constant in the body frame rather than in the world frame between two consecutive IMU measurements. Even though higher-order numerical methods \cite{munthe1999high}, \cite{andrle2013geometric} could be used  to improve the IMU integration method in \cite{7557075}, the complexity of the high-order solution could be an issue. On the other hand, it is very difficult to  take  into account the IMU biases and to compute the covariance of the high-order IMU integration.  It is thus necessary to set up a new IMU integration model which based on more realistic assumptions.

In this paper, a novel IMU Preintegration model is introduced by using switched linear systems \cite{li2005switched}. The new model assumes that {\it both the linear acceleration and the angular velocity in the body frame are constant between two consecutive IMU measurements}. The assumption that the linear acceleration in the body frame is constant between two consecutive IMU measurements is reasonable because external forces, such as thrust, are of fixed orientation in the body frame \cite{Hamel2002visual}. A simple discrete linear system is first developed to represent the kinematic dynamics between two successive IMU measurements. Since both the linear acceleration and the angular velocity between two successive visual frames are usually time-varying, the overall system between two successive visual frames is a switched linear system \cite{li2005switched}. The IMU preintegration between two video frames is then obtained by integrating the switched linear system using all the IMU measurements between the two video frames. The proposed approach provides an analytic solution to the motion integration rather than an approximation solution as in \cite{7557075}. Thus, it presents a more accurate
motion estimation regardless of the value of angular velocity between two successive IMU measurements. Experimental results show that the proposed model outperforms the classical model in \cite{7557075}. Overall, the main contribution of this paper is a novel IMU preintegration model which gives a better motion accuracy because the kinematics between two video frames is analyzed using  switched linear systems. Since the proposed model can be applied to improve visual odmetry   and Lidar odometry, it is useful for localization of high speed autonomous vehicles in GPS denied environments.

This paper is organized as follow. The classical IMU integration model is presented in section \ref{TRO2017}. A new model is introduced in section \ref{newmodels}. The performance of the new model is evaluated in section \ref{result} using both real datasets. Finally, section \ref{conculsionsession} concludes this paper.

\section{Related Works on IMU Preintegration}
\label{TRO2017}

In this section, the detail on the IMU preintegration model in \cite{7557075} is provided. An IMU provides both linear acceleration and angular velocity in the body frame in the form of a 3-axis accelerometer and 3-axis gyroscope data denoted by $a_B$ and $\omega _B$. The body frame is expressed by the suffix B notation while the world frame is expressed by the suffix W notation. The linear acceleration vector and the angular velocity vector are transformed into the world frame using the rotational matrix from the body frame to the world frame. The gravity vector in the world frame is denoted by $g_W$.

Let $P_W=\left[\begin{array}{lll}
P_1 & P_2 & P_3\end{array}\right]^T$ and $V_W=\left[\begin{array}{lll} V_1 & V_2 & V_3\end{array}\right]^T$  be the position of the camera center and the linear velocity of the agent having the camera in the world reference frame and $R_{BW}$ be the
rotation matrix from the body frame to the world reference one.
The IMU integration can be obtained by introducing the IMU kinematic model as \cite{7557075, mur2017visual}:
\begin{equation}
\left\{\begin{array}{lcr}
\dot{P}_W = V_W\\
\dot{V}_W = a_W \\
\dot{R}_{BW} = R_{BW}\omega_B^{\wedge}
\end{array}\right.,
\label{eu1}
\end{equation}
\begin{flushleft} where $\omega_B=\left[\begin{array}{lll} \omega_1 & \omega_2 & \omega_3\end{array}\right]^T$  is the angular velocity of the agent measured by a gyroscope in the body frame. The matrix $\omega_B^{\wedge}$ is
\end{flushleft}
\begin{equation}
\omega_B^{\wedge}=\left[\begin{array}{lcr}
0 & -\omega_3 & \omega_2\\
\omega_3 & 0 & -\omega_1\\
-\omega_2 & \omega_1 & 0\\
\end{array}
\right].
\label{eu2}
\end{equation}

It can be easily derived that
\begin{equation}
\left\{\begin{array}{l}
(\omega_B^{\wedge})^2=\left[\begin{array}{lcr}
-\omega_2^2-\omega_3^2 & \omega_1\omega_2 & \omega_1\omega_3\\
\omega_1\omega_2 & -\omega_1^2-\omega_3^2 & \omega_2\omega_3\\
\omega_1\omega_3 & \omega_2\omega_3 & -\omega_1^2-\omega_2^2\\
\end{array}
\right]\\
(\omega_B^{\wedge})^3=-\|\omega_B\|^2\omega_B^{\wedge}\\
\end{array}
\right.,
\label{eu222}
\end{equation}
where $\|\omega_B\|^2$ is $(\omega_1^2+\omega_2^2+\omega_3^2)$.

\begin{flushleft} The state estimate at $t+{\bigtriangleup}t$ is obtained by the integration of kinematic model (\ref{eu1}) from $t$ to  $t+{\bigtriangleup}t$ as
\end{flushleft}
\begin{equation}
\left\{\begin{array}{lcr}
P_W(t+{\Delta}t)=\displaystyle P_W(t)+ \int_t^{t+{\Delta}t}\limits V_W(\tau)d\tau + \int\int_t^{t+{\Delta}t}\limits a_W(\tau)d\tau^2\\
V_W(t+{\Delta}t)=\displaystyle V_W(t)+ \int_t^{t+{\Delta}t}\limits a_W(\tau)d\tau\\
R_{BW}(t+{\Delta}t)=\displaystyle R_{BW}(t) exp(\int_t^{t+{\Delta}t}\limits \omega_B(\tau)d\tau)\\
\end{array}
\right..
\label{eu3}
\end{equation}

Generally, equation (\ref{eu3}) does not have an exact solution for measurements with time varying linear acceleration and angular velocity. By assuming that  the linear acceleration in the world frame $a_W$ and the angular velocity in the body frame $\omega_B$ are constant in the time interval [$t$, $t+{\Delta}t$] \cite{7557075}, the kinematic model can be expressed by
\begin{equation}
\left\{\begin{array}{lcr}
P_W(t+{\Delta}t)=\displaystyle P_W(t)+ V_W \Delta t + \frac{1}{2} a_W \Delta t^2\\
V_W(t+{\Delta}t)=\displaystyle V_W(t)+ a_W \Delta t\\
R_{BW}(t+{\Delta}t)=\displaystyle R_{BW}(t) E(\theta_B(t))\\
\end{array}
\right.,
\label{eu4}
\end{equation}
where   $\theta_B(t)$ is $\omega_B(t)\Delta t$, and the matrix $E(\theta_B(t))$ is defined as in   Table \ref{matrices123}. When the values of $\omega_i(i=1,2,3)$ are zero's, the matrix $E(\theta)$ becomes the $3\times 3$ identity matrix $I$.

\begin{table}[!h]
\caption{Coefficients of the matrices $E(\theta)$, $\Gamma(\theta)$ and $\Lambda(\theta)$}
\label{table}
\vspace{-3mm}
\begin{center}
\tabcolsep 0.06in
\renewcommand{\arraystretch}{1.5}
\begin{tabular}{|c|c|c|c|}
\hline
Matrix  & Coefficient of $I$ & Coefficient of $\theta^{\wedge}$  & Coefficient of $(\theta^{\wedge})^2$  \\
\hline
$E(\theta)$ & 1 & $\frac{h_1(\|\theta\|)}{\|\theta\|}$ & $\frac{h_2(\|\theta\|)}{\|\theta\|^2}$\\\hline
$\Gamma(\theta)$ & 1 & $\frac{h_2(\|\theta\|)}{\|\theta\|^2}$ & $\frac{h_3(\|\theta\|)}{\|\theta\|^3}$ \\\hline
 $\Lambda(\theta)$ & $\frac{1}{2}$ & $\frac{h_3(\|\theta\|)}{\|\theta\|^3}$ & $\frac{h_4(\|\theta\|)}{2\|\theta\|^4}$ \\\hline
\end{tabular}
\end{center}
\label{matrices123}
\end{table}

\begin{flushleft}The functions $h_i(z)(i=1,2,3,4)$ in Table \ref{matrices123}  are defined as
\end{flushleft}
\begin{equation}
\left\{\begin{array}{l}
 h_1(z)=\sin z\\
h_2(z) = 1-\cos z\\
h_3(z) = z-\sin z\\
h_4(z) = 2\cos z-2+z^2\\
\end{array}
\right..
\end{equation}

\begin{flushleft}It can be easily verified that
\end{flushleft}
\begin{equation}
\left\{\begin{array}{l}
\theta^{\wedge}\Gamma(\theta)+I=E(\theta)\\
\theta^{\wedge}\Lambda(\theta)+I=\Gamma(\theta)\\
\end{array}
\right..
\end{equation}

\begin{flushleft} Notice that $a_W$ can be expressed by:
\end{flushleft}
\begin{equation}
a_W(t)=R_{BW}(t)a_B+g_W,
\label{eu5}
\end{equation}
where $a_B=\left[\begin{array}{lll}a_1 & a_2 & a_3\end{array}\right]^T$  is the linear acceleration in the body  frame that is measured with an accelerometer,  and its value is \cite{Hamel2002visual}
\begin{equation}
\label{a_B}
a_B = \frac{u_BF_B}{m},
\end{equation}
$u_B$ is the magnitude of thrust applied to the agent, $F_B$ is a constant unit vector in the body frame representing the fixed orientation of the thruster, and $m$ is the mass of the agent. $g_W=\left[\begin{array}{lll}0 & 0 & -9.81\end{array}\right]^T$  with $g_W$ being the gravity acceleration in the world frame.

 Substituting (\ref{eu5}) in (\ref{eu4}), the IMU kinematic model becomes
\begin{equation}
\left\{\begin{array}{lcr}
P_W(t+{\Delta}t)=\displaystyle P_W(t)+ V_W \Delta t + \frac{1}{2} g_W \Delta t^2 +\frac{1}{2} R_{BW}(t) a_B \Delta t^2\\
V_W(t+{\Delta}t)=\displaystyle V_W(t)+ g_W \Delta t + R_{BW}(t) a_B \Delta t\\
R_{BW}(t+{\Delta}t)=\displaystyle R_{BW}(t) E(\theta_B(t))\\
\end{array}
\right..
\label{eu6}
\end{equation}

The principle of IMU preintegration theory is to integrate all IMU measurements between two visual keyframes in a single compound measurement. The IMU and camera can provide  measurements synchronously at the discrete time $k$. Suppose that two video frames are captured at the time instances $k=i$ and $k=j$. The IMU integrated model can be obtained by integration of (\ref{eu6}) between the two frames  as shown in the following equation:
\begin{equation}
\left\{\begin{array}{lcr}
P_W(j)=P_W(i)+\Theta(i,j)+R_{BW}(i)\zeta_1(i,j)\\
V_W(j)=V_W(i)+g_W{\displaystyle\sum_{k=i}^{j-1}}\Delta t+R_{BW}(i)\mu_1(i, j)\\
 R_{BW}(j)=R_{BW}(i)F(i,j)\\
 \end{array}
\right.,
\label{eu77}
\end{equation}
where the vector $\Theta(i,j)$ and the matrix $F(i,j)$ are computed as
  \begin{equation}
\left\{\begin{array}{ll}
\Theta(i,j) = V_W(i){\displaystyle\sum_{k=i}^{j-1}}\Delta t+\frac{g_W}{2}({\displaystyle\sum_{k=i}^{j-1}}\Delta t)^2\\
F(i,j)={\displaystyle \prod_{k=i}^{j-1}}E(\theta_B(k))\\
\end{array}
\right.,
\label{eu777}
\end{equation}
and the vectors   $\zeta_1(i, j)$ and $\mu_1(i, j)$  are given as
  \begin{equation}
\left\{\begin{array}{ll}
 \zeta_1(i,j)= {\displaystyle\sum_{k=i}^{j-1}}(\frac{1}{2}F(i,k)a_B(k)\Delta t^2+\mu_1(i, k)\Delta t)\\
\mu_1(i, j)={\displaystyle\sum_{k=i}^{j-1}}F(i,k)a_B(k)\Delta t\\
\end{array}
\right..
\label{zeta1mu1}
\end{equation}

\begin{flushleft} The translation matrix between the two video frames $i$ and $j$, $T(i,j)$ is finally computed as
\end{flushleft}
\begin{equation}
\label{tranformationmatrix}
T(i,j)=\left[\begin{array}{ll}
F^T(i, j) & t(i,j)\\
0 & 1\\
\end{array}
\right],
\end{equation}
where the  translation vector $t(i,j)$ is $(-F^T(i,j)(\zeta_1(i,j)+R_{WB}(i)\Theta(i,j)))$.

The equations (\ref{eu77}) and (\ref{eu777}) can be regarded as the classical IMU integration model. It can be shown from the equations (\ref{eu5}) and (\ref{a_B}) that the linear acceleration can be assumed to be constant in the world frame if the value of $\omega_B$ is zero between two successive IMU measurements. As such, the agent orientation $R_{BW}(t)$ is constant in the time interval $[t, t+\Delta t]$. All the integrations in the equation (\ref{eu6}) are accurate. It was pointed out in \cite{7557075} that this assumption is usually true if the the sampling rate of IMU is high. But it could be an issue if 1) the sampling rate is slow or 2) the value of $\omega_B$ is not zero between two successive IMU measurements at a high sampling rate. According to the equation (\ref{a_B}), {\it $a_B$ intuitively has more chance than $a_W$ to be  constant in the  small interval $[t, t+\Delta t]$}.

\section{A New IMU Preintegration Model Using Switched Linear Systems}
\label{newmodels}

In this section, a new IMU integration model is  derived under an assumption that {\it both the linear acceleration  and the angular velocity in the body frame  are constant between two successive IMU measurements}.

Let $R_{WB}=\left[\begin{array}{lll} R_1 & R_2 & R_3\end{array}\right]$ be the
rotation matrix from the world reference frame to the body frame. A new kinematics model will be built up using the new state vector whereby new state vector is introduced as $\left[\begin{array}{lllll}P_W^T & V_W^T & R_1^T & R_2^T& R_3^T\end{array}\right]^T$. Let $I$ be  a $3\times 3$ identity matrix. Using the equation $R_{WB}R_{BW}=I$, it can be derived from the equations (\ref{eu1}) and (\ref{eu5}) that
 \begin{equation}
 \left[\begin{array}{l}
 \dot{P}_W\\
 \dot{V}_W\\
 \dot{R}_1\\
 \dot{R}_2\\
 \dot{R}_3\\
 \end{array}
 \right]=\left[\begin{array}{c}
 V_W\\
 R_{BW}a_B+g_W\\
 -\omega_B^{\wedge}R_1\\
 -\omega_B^{\wedge}R_2\\
 -\omega_B^{\wedge}R_3\\
 \end{array}
 \right],
 \end{equation}
\begin{flushleft} Using the equality that $R_{BW}=R_{WB}^T$ and the matrix theory \cite{horn2012matrix}, it can be derived that
\end{flushleft}
\begin{eqnarray}
R_{BW} a_B=\left[\begin{array}{l}R_1^Ta_B\\
R_2^Ta_B\\
R_3^Ta_B\\\end{array}\right]=\left[\begin{array}{l}a^T_BR_1\\
a^T_BR_2\\
a^T_BR_3\\\end{array}\right].
\end{eqnarray}
\begin{flushleft} It follows that
\end{flushleft}
\begin{equation}
 \label{agentdynamics}
 \left[\begin{array}{l}
 \dot{P}_W\\
 \dot{V}_W\\
 \dot{R}_1\\
 \dot{R}_2\\
 \dot{R}_3\\
 \end{array}
 \right]=A(1,a_B,\omega_B)\left[\begin{array}{l}
 P_W\\
V_W\\
R_1\\
R_2\\
R_3\\
 \end{array}
 \right]+\left[\begin{array}{l}0\\g_W\\ 0\\ 0 \\ 0\\
 \end{array}
 \right],
 \end{equation}
\begin{flushleft}where the matrix $A(s,a,\omega)$ is given by
\end{flushleft}
\begin{eqnarray}
&&\hspace{-9mm}A(s,a,\omega) = \left[\begin{array}{lllll}
0 & sI & 0 & 0 & 0\\
0& 0 & \hat{A}_1(a) & \hat{A}_2(a) & \hat{A}_3(a)\\
0&  0 & -\omega^{\wedge}  & 0 & 0\\
0&  0 & 0 & -\omega^{\wedge} &  0\\
0&  0 & 0 & 0&  -\omega^{\wedge}  \\
 \end{array}
 \right],\end{eqnarray}
  and the matrices $\hat{A}_i(a)(1\leq i\leq 3)$ are
\begin{equation}
\left\{\begin{array}{lll}
\hat{A}_1(a) = \left[\begin{array}{lll}
 a & 0 & 0\\
  \end{array}
 \right]^T\\
\hat{A}_2(a) = \left[\begin{array}{lll}
0 & a  & 0\\
 \end{array}
 \right]^T\\
\hat{A}_3(a) = \left[\begin{array}{lll}
 0 & 0 & a \\
  \end{array}
 \right]^T\\
 \end{array}
\right..
\end{equation}

The new kinematics model (\ref{agentdynamics}) is a time varying linear system.  Assuming that the values of $a_B$ and $\omega_B$ are fixed in the small interval $[k\Delta t, (k+1)\Delta t]$, the matrix $A(1,a_B,\omega_B)$ is a constant matrix in the interval. Subsequently, the system in the equation (\ref{agentdynamics}) is a linear time invariant system in the interval. Using the linear control system theory \cite{kailath1980linear}, a discrete model can be derived from the  equation (\ref{agentdynamics}) as \cite{horn2012matrix,kailath1980linear}
\begin{equation}
 \label{discretemodel}
 \left[\begin{array}{l}
 P_W(k+1)\\
V_W(k+1)\\
R_1(k+1)\\
R_2(k+1)\\
R_3(k+1)\\
 \end{array}
 \right]=e^{A(\Delta t, \Delta t a_B,\theta_B)}\left[\begin{array}{l}
 P_W(k)\\
V_W(k)\\
R_1(k)\\
R_2(k)\\
R_3(k)\\
 \end{array}
 \right]+\left[\begin{array}{l}
 g_W\frac{\Delta t^2}{2}\\
 g_W\Delta t \\
 0\\
 0\\
 0\\
 \end{array}
 \right],
 \end{equation}
\begin{flushleft} where the matrix $e^{A(s, a,\theta)}$ is  given as
\end{flushleft}
\vspace*{1mm}

\begin{equation}
\left[\begin{array}{lllll}
I & sI & s\hat{A}_1(a)\Lambda(-\theta) & s\hat{A}_2(a)\Lambda(-\theta) & s\hat{A}_3(a)\Lambda(-\theta)\\
0 & I & \hat{A}_1(a)\Gamma(-\theta) & \hat{A}_2(a)\Gamma(-\theta) & \hat{A}_3(a)\Gamma(-\theta)\\
0 &  0 & E(-\theta)  & 0 & 0\\
0 &  0 & 0 & E(-\theta) &  0\\
0 &  0 & 0 & 0&  E(-\theta)  \\
\end{array}
\right],
\end{equation}
and the matrices $\Gamma(\theta)$ and $\Lambda(\theta)$    are defined as in  Table \ref{matrices123}.

Using the following equation
\begin{equation}
\left\{\begin{array}{l}
a_B^T(k)\Lambda(-\theta_B(k))R_i(k)=R_i^T(k)\Lambda(\theta_B(k))a_B(k)\\
a_B^T(k)\Gamma(-\theta_B(k))R_i(k)=R_i^T(k)\Gamma(\theta_B(k))a_B(k)\\
(E(-\theta_B(k))R_i)^T=R_i^TE(\theta_B(k))
\end{array}
\right.,
\end{equation}
it can be shown that the equation (\ref{discretemodel}) is equivalent to
\begin{equation}
\left\{\begin{array}{lccr}
P_W(k+1)=\displaystyle P_W(k)+ V_W(k) \Delta t\\
 \hspace{17mm}+ (\frac{1}{2} g_W+ R_{BW}(k) \Lambda(\theta_B(k))a_B(k)) \Delta t^2\\
V_W(k+1)=\displaystyle V_W(k)+ (g_W  + R_{BW}(k) \Gamma(\theta_B(k))a_B(k)) \Delta t\\
R_{BW}(k+1)=\displaystyle  R_{BW}(k)E(\theta_B(k))\\
\end{array}
\right..
\label{eu666}
\end{equation}

 Similar to \cite{7557075}, we also assume that two video frames are captured at the time instances $k=i$ and $k=j$.
  Defining a matrix $\hat{A}(k)$ as
 \begin{equation}
 \hat{A}(k)=\exp^{A(\Delta t,\Delta ta_B(k),\theta_B(k))},
 \end{equation}
  and a vector $\hat{b}$ as
 \begin{equation}
 \hat{b}=\left[\begin{array}{lllll}
 g_W\frac{\Delta t^2}{2} &
 g_W\Delta t &
 0 &
 0&
 0
 \end{array}
 \right]^T,
 \end{equation}
the kinematics of the agent between the two video frames can be represented by the following time variant system:
\begin{equation}
\label{switchedsystem}
\left[\begin{array}{l}
 P_W(k+1)\\
V_W(k+1)\\
R_1(k+1)\\
R_2(k+1)\\
R_3(k+1)\\
 \end{array}
 \right]=\hat{A}(k)\left[\begin{array}{l}
 P_W(k)\\
V_W(k)\\
R_1(k)\\
R_2(k)\\
R_3(k)\\
 \end{array}
 \right]+\hat{b},
\end{equation}
where $i\leq k\leq j$.

The values of $a_B$ and $\omega_B$ may be changed in the next small time interval $[(k+1)\Delta t, (k+2)\Delta t]$. This implies that the matrix $\hat{A}(k)$  could be different from the matrix $\hat{A}(k+1)$. Thus, the time variant system in the equation (\ref{switchedsystem}) is a switched linear system \cite{li2005switched}. The discrete dynamic model in the equation (\ref{switchedsystem}) will be used to integrate all the IMU data between the two video frames as follows:
\begin{equation}
\left\{\begin{array}{lcr}
P_W(j)=P_W(i)+\Theta(i,j)+ R_{BW}(i)\zeta_2(i,j)\\
V_W(j)=V_W(i)+g_W{\displaystyle\sum_{k=i}^{j-1}}\Delta t+R_{BW}(i)\mu_2(i, j)\\
 R_{BW}(j)=R_{BW}(i)F(i,j)\\
 \end{array}
\right.,
\label{eq1}
\end{equation}
where the vectors  $\zeta_2(i, j)$ and $\mu_2(i, j)$ a are computed as
  \begin{equation}
  \label{zeta2mu2}
\left\{\begin{array}{lll}
 \zeta_2(i,j)= {\displaystyle\sum_{k=i}^{j-1}}(F(i,k)\Lambda(\theta_B(k))a_B(k)\Delta t^2+\mu_2(i, k)\Delta t)\\
\mu_2(i, j)={\displaystyle\sum_{k=i}^{j-1}}F(i,k)\Gamma(\theta_B(k))a_B(k)\Delta t\\
\end{array}
\right..
\end{equation}

\vspace*{0.2mm}

The rotation matrix for the proposed model is $F^T(i,j)$, and the  translation vector $t(i,j)$ for the proposed model is $(-F^T(i,j)(\zeta_2(i,j)+R_{WB}(i)\Theta(i,j)))$.

\section{Experimental Results}
\label{result}

In this section, the proposed model is compared with the classical model in \cite{7557075} using real public datasets. These datasets include different conditions in flight dynamics, texture and illumination. For the fair comparison, we have implemented both models on VINS\_MONO benchmark which it is an open source code \cite{qin2017vins}. The evaluation has been done using high sampling IMU 200Hz and all key frames have been selected at 10Hz. The loop closure has also been enabled and used in this evaluation test. The estimation output from VINS\_MONO benchmark has a different reference from the ground truth reference. Thus, the estimation raw data has been aligned first with the ground truth reference using the trajectory evaluation toolbox in \cite{zhang2018tutorial}.  It is worth noting that the alignment process has been done using the full trajectory states.

The EuRoC dataset has been used in this test where it provides the measurement using a micro aerial vehicle (MAV) in different places with different conditions in flight dynamics, texture and illumination. The dataset has been classified into different sequences as easy, medium and difficult. The measurements have been recorded at 200Hz for IMU and 20Hz for images. The dataset has been synchronized, corrected and matched with the IMU measurements output \cite{Burri25012016}. As we evaluate both models on monocular visual inertial odometry, the images from the left camera have been used only.

\subsection{Difficult Conditions Sequences}
\label{difficult}

\begin{figure}[!t]
\centering
\vspace{0.2cm}
\includegraphics[width=2.8in,height=2.8in,trim={2.0cm 0.5cm 2.0cm 0.5cm}]{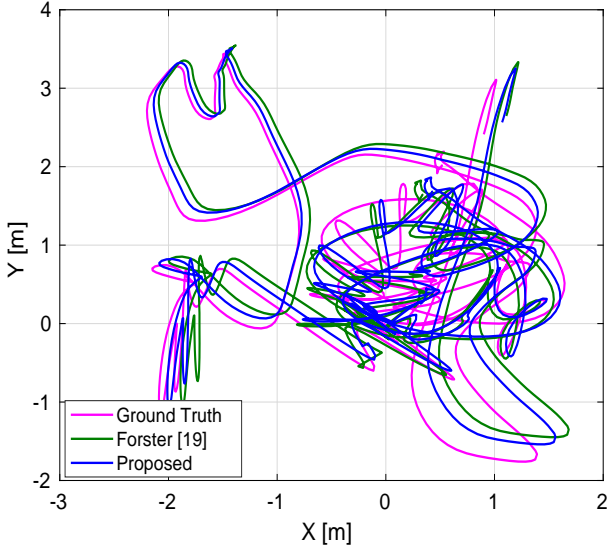}
\caption{Estimation trajectory against ground-truth for the proposed model and the model in \cite{7557075}}
\label{Trajectory}
\end{figure}

\begin{figure}[!htb]
	\centering{
\subfigure[Overall position error]{\includegraphics[width=2.5in,height=1.5in,trim={4cm 0cm 4cm 0cm}]{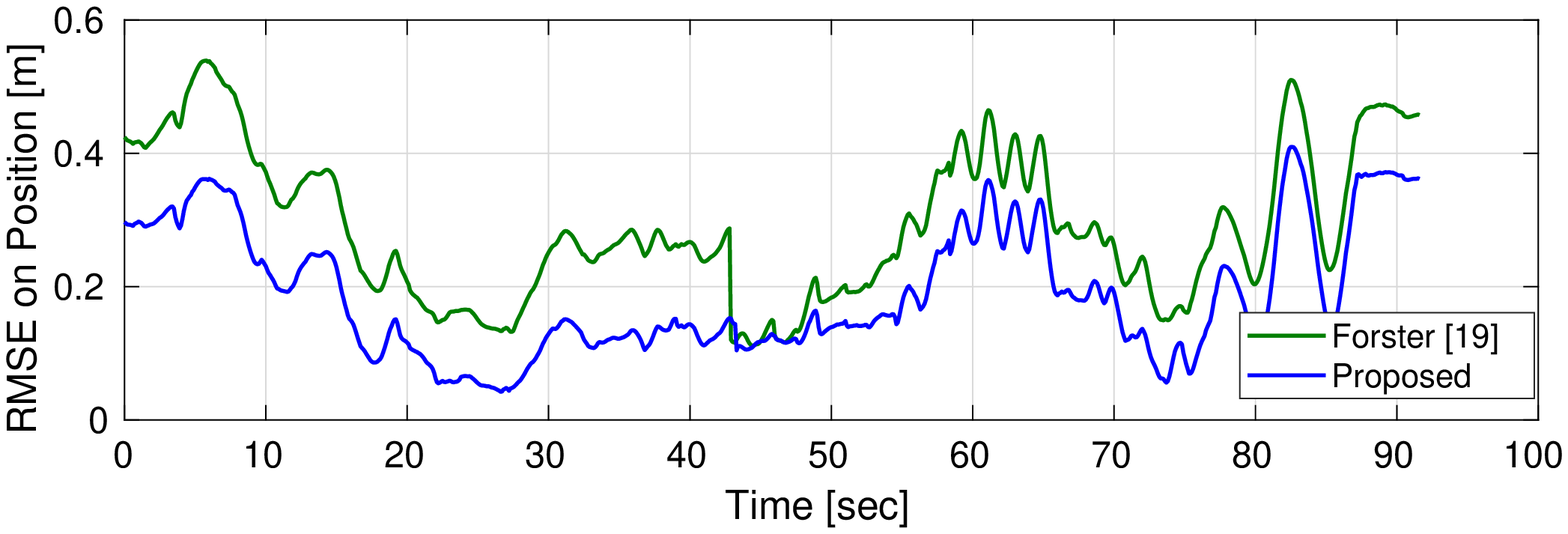}}
\subfigure[Position error at X direction]{\includegraphics[width=2.5in,height=1.5in,trim={4cm 0cm 4cm 0cm}]{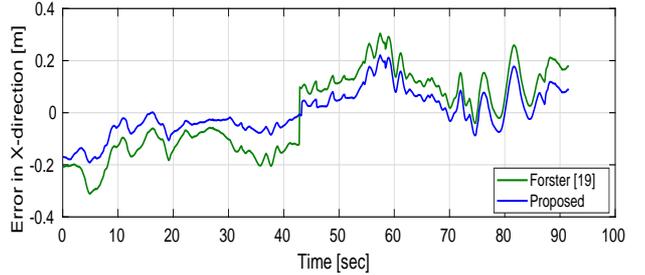}}
\subfigure[Position error at Y direction]{\includegraphics[width=2.5in,height=1.5in,trim={4cm 0cm 4cm 0cm}]{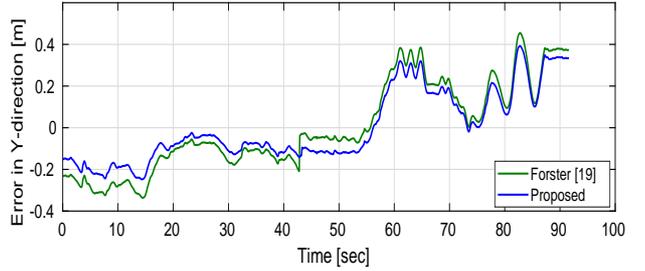}}
\subfigure[Position error at Z direction]{\includegraphics[width=2.5in,height=1.5in,trim={4cm 0cm 4cm 0cm}]{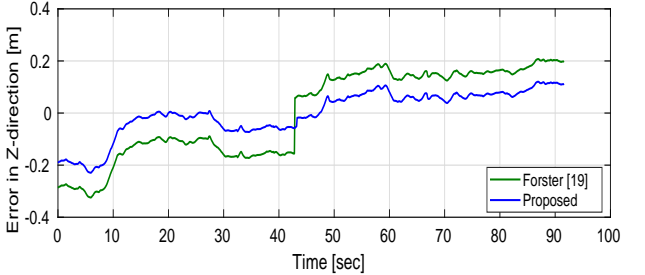}}
}
\caption{RMSE and position analysis on V1\_03\_difficult Sequence for the proposed model and the model in \cite{7557075}.}
\label{analysis}
\end{figure}

In this subsection, the comparison between the state of the art and the proposed approach has been done on difficult conditions sequences which presents the most challenging conditions in the terms of lighting and flights dynamics and texture-less areas. Accordingly, a V1\_03\_difficult sequence has been selected to perform this analysis. Fig. \ref{Trajectory} shows the estimated trajectories of the state of the art and proposed approaches compared to the ground-truth. In addition, the accuracy of the proposed approach is evaluated against model in \cite{7557075}. The RMSE and position errors in x-direction, y-direction and z-direction have been illustrated in Fig. \ref{analysis}. The proposed model estimates the position more accurately at x, y and z directions whereby the proposed model outperforms the model in \cite{7557075} by 31.84$\%$ over the full path of V1\_03\_difficult sequence. Not only the proposed model outperforms the classical model on  V1\_03\_difficult sequence, but also it outperforms the classical model on both sequences V2\_03\_difficult and MH\_05\_difficult by 19.89$\%$ and 11.11$\%$, respectively. The reason that, the proposed can handle the dramatic dynamic changes and gives a reliable estimation over the classical one, the proposed model takes into consideration the dynamic changes between measurements. The metric RMSE errors of both sequences have been illustrated at Table \ref{RMSEposition}.

\subsection{Comparison on Different Sequences}
\label{different}

In this subsection, the proposed model has been evaluated using different sequences to test the accuracy of the proposed model over the classical one in different environment. Table \ref{RMSEposition} shows the root mean square error (RMSE) over the full path of all sequences which have been used in this evaluation.  The proposed model  outperforms the classical model on MH\_02\_easy and V2\_01\_easy sequences by 3.89$\%$ and 2.32$\%$, respectively. On the other hand, the classical model outperforms the proposed model on V1\_01\_easy sequence but with a very small difference by 1.3 $mm$ only. When the condition of motion becomes aggressive and significant changing in illumination with texture-less area, the proposed model can gives much better estimation. This implies that the proposed model achieves higher accuracy than the state of the art especially in these cases when high dynamic changes in angular velocity or significant illumination changing environment with texture-less area.


\begin{table}[!htb]
\caption{RMSE On Position For Different Sequences}
\label{table}
\vspace{-3mm}
\begin{center}
\tabcolsep 0.06in
\renewcommand{\arraystretch}{1.5}
\begin{tabular}{|c|c|c|c|}
\hline
Sequence &  Forster \cite{7557075} [cm]& Proposed [cm]& Percentage [$\%$] \\\hline
MH\_02\_easy &  16.70 & \textbf{16.05} & 3.89\\\hline
MH\_05\_difficult &  30.32 & \textbf{26.95} & 11.11\\\hline
V1\_01\_easy &  \textbf{10.16} & 10.29 & -1.28\\\hline
V1\_03\_difficult &  31.12 & \textbf{21.21} & 31.84\\\hline
V2\_01\_easy &  7.33 & \textbf{7.16} & 2.32\\\hline
V2\_03\_difficult &  28.10 & \textbf{22.51} & 19.89\\\hline
\end{tabular}
\end{center}
\label{RMSEposition}
\end{table}

\section{CONCLUSIONS}
\label{conculsionsession}

A new IMU preintegration model has been proposed with formulation using switched linear systems. The proposed model assumes constant linear acceleration in the body frame. Doing so allow our proposed model to be closer to real world condition. Our proposed model is compared with state-of-the-art model on real datasets  with different conditions in flight dynamics, texture and illumination. Experimental results show that the proposed model outperforms the state-of-the-art model.

There are many applications of the proposed model. For example, it can utilized to improve visual odmetry  \cite{valenti2018enabling,7557075} and Lidar odometry  \cite{javanmardi2017autonomous,zhang2014loam}. These applications are important for high speed autonomous vehicles. For a potential contact-based inspection operation within UAV framework \cite{mahmoud2014linear,kocer2019model}, it can be a significant stage to leverage such algorithms in real time. All these works will be studied in our future research.

\addtolength{\textheight}{-1cm}   





\section*{ACKNOWLEDGMENT}

This research is partially supported by SERC grant No. 162 25 00036 from the National Robotics Programme (NRP) and No. A1715a0064 from Aerospace Program, Singapore.


\bibliography{IEEEabrv,references}

\begin{thebibliography}{10}
\providecommand{\url}[1]{#1}
\csname url@samestyle\endcsname
\providecommand{\newblock}{\relax}
\providecommand{\bibinfo}[2]{#2}
\providecommand{\BIBentrySTDinterwordspacing}{\spaceskip=0pt\relax}
\providecommand{\BIBentryALTinterwordstretchfactor}{4}
\providecommand{\BIBentryALTinterwordspacing}{\spaceskip=\fontdimen2\font plus
\BIBentryALTinterwordstretchfactor\fontdimen3\font minus
  \fontdimen4\font\relax}
\providecommand{\BIBforeignlanguage}[2]{{%
\expandafter\ifx\csname l@#1\endcsname\relax
\typeout{** WARNING: IEEEtran.bst: No hyphenation pattern has been}%
\typeout{** loaded for the language `#1'. Using the pattern for}%
\typeout{** the default language instead.}%
\else
\language=\csname l@#1\endcsname
\fi
#2}}
\providecommand{\BIBdecl}{\relax}
\BIBdecl

\bibitem{javanmardi2017autonomous}
E.~Javanmardi, M.~Javanmardi, Y.~Gu, and S.~Kamijo, ``Autonomous vehicle
  self-localization based on probabilistic planar surface map and multi-channel
  lidar in urban area,'' in \emph{2017 IEEE 20th International Conference on
  Intelligent Transportation Systems (ITSC)}.\hskip 1em plus 0.5em minus
  0.4em\relax IEEE, 2017, pp. 1--8.

\bibitem{murali2017utilizing}
V.~Murali, H.-P. Chiu, S.~Samarasekera, and R.~T. Kumar, ``Utilizing semantic
  visual landmarks for precise vehicle navigation,'' in \emph{2017 IEEE 20th
  International Conference on Intelligent Transportation Systems (ITSC)}.\hskip
  1em plus 0.5em minus 0.4em\relax IEEE, 2017, pp. 1--8.

\bibitem{valenti2018enabling}
F.~Valenti, D.~Giaquinto, L.~Musto, A.~Zinelli, M.~Bertozzi, and A.~Broggi,
  ``Enabling computer vision-based autonomous navigation for unmanned aerial
  vehicles in cluttered gps-denied environments,'' in \emph{2018 21st
  International Conference on Intelligent Transportation Systems (ITSC)}.\hskip
  1em plus 0.5em minus 0.4em\relax IEEE, 2018, pp. 3886--3891.

\bibitem{trawny2007vision}
N.~Trawny, A.~I. Mourikis, S.~I. Roumeliotis, A.~E. Johnson, and J.~F.
  Montgomery, ``Vision-aided inertial navigation for pin-point landing using
  observations of mapped landmarks,'' \emph{Journal of Field Robotics},
  vol.~24, no.~5, pp. 357--378, 2007.

\bibitem{li2019asymptotic}
Z.~Li, W.~Gao, C.~Goh, M.~Yuan, E.~K. Teoh, and Q.~Ren, ``Asymptotic
  stabilization of nonholonomic robots leveraging singularity,'' \emph{IEEE
  Robotics and Automation Letters}, vol.~4, no.~1, pp. 41--48, 2019.

\bibitem{jones2011visual}
E.~S. Jones and S.~Soatto, ``Visual-inertial navigation, mapping and
  localization: A scalable real-time causal approach,'' \emph{The International
  Journal of Robotics Research}, vol.~30, no.~4, pp. 407--430, 2011.

\bibitem{abdi2016pose}
G.~Abdi, F.~Samadzadegan, and F.~Kurz, ``Pose estimation of unmanned aerial
  vehicles based on a vision-aided multi-sensor fusion,''
  \emph{ISPRS-International Archives of the Photogrammetry, Remote Sensing and
  Spatial Information Sciences}, pp. 193--199, 2016.

\bibitem{hemann2016long}
G.~Hemann, S.~Singh, and M.~Kaess, ``Long-range gps-denied aerial inertial
  navigation with lidar localization,'' in \emph{Intelligent Robots and Systems
  (IROS), 2016 IEEE/RSJ International Conference on}.\hskip 1em plus 0.5em
  minus 0.4em\relax IEEE, 2016, pp. 1659--1666.

\bibitem{zhang2017low}
J.~Zhang and S.~Singh, ``Low-drift and real-time lidar odometry and mapping,''
  \emph{Autonomous Robots}, vol.~41, no.~2, pp. 401--416, 2017.

\bibitem{kumar2017lidar}
G.~A. Kumar, A.~K. Patil, R.~Patil, S.~S. Park, and Y.~H. Chai, ``A lidar and
  imu integrated indoor navigation system for uavs and its application in
  real-time pipeline classification,'' \emph{Sensors}, vol.~17, no.~6, p. 1268,
  2017.

\bibitem{5427255}
A.~Nemra and N.~Aouf, ``Robust ins/gps sensor fusion for uav localization using
  sdre nonlinear filtering,'' \emph{IEEE Sensors Journal}, vol.~10, no.~4, pp.
  789--798, April 2010.

\bibitem{perez2016fusion}
G.~Perez-Paina, C.~Paz, M.~Kulich, M.~Saska, and G.~Aragu{\'a}s, ``Fusion of
  monocular visual-inertial measurements for three dimensional pose
  estimation,'' in \emph{International Workshop on Modelling and Simulation for
  Autonomous Systems}.\hskip 1em plus 0.5em minus 0.4em\relax Springer, 2016,
  pp. 242--260.

\bibitem{grisetti2010tutorial}
G.~Grisetti, R.~Kummerle, C.~Stachniss, and W.~Burgard, ``A tutorial on
  graph-based slam,'' \emph{IEEE Intelligent Transportation Systems Magazine},
  vol.~2, no.~4, pp. 31--43, 2010.

\bibitem{shen2015tightly}
S.~Shen, N.~Michael, and V.~Kumar, ``Tightly-coupled monocular visual-inertial
  fusion for autonomous flight of rotorcraft mavs,'' in \emph{2015 IEEE
  International Conference on Robotics and Automation (ICRA)}.\hskip 1em plus
  0.5em minus 0.4em\relax IEEE, 2015, pp. 5303--5310.

\bibitem{yang2017monocular}
Z.~Yang and S.~Shen, ``Monocular visual--inertial state estimation with online
  initialization and camera--imu extrinsic calibration,'' \emph{IEEE
  Transactions on Automation Science and Engineering}, vol.~14, no.~1, pp.
  39--51, 2017.

\bibitem{mourikis2007multi}
A.~I. Mourikis and S.~I. Roumeliotis, ``A multi-state constraint kalman filter
  for vision-aided inertial navigation,'' in \emph{Proceedings 2007 IEEE
  International Conference on Robotics and Automation}.\hskip 1em plus 0.5em
  minus 0.4em\relax IEEE, 2007, pp. 3565--3572.

\bibitem{brossard2018invariant}
M.~Brossard, S.~Bonnabel, and A.~Barrau, ``Invariant kalman filtering for
  visual inertial slam,'' in \emph{2018 21st International Conference on
  Information Fusion (FUSION)}.\hskip 1em plus 0.5em minus 0.4em\relax IEEE,
  2018, pp. 2021--2028.

\bibitem{6092505}
T.~Lupton and S.~Sukkarieh, ``Visual-inertial-aided navigation for high-dynamic
  motion in built environments without initial conditions,'' \emph{IEEE
  Transactions on Robotics}, vol.~28, no.~1, pp. 61--76, Feb 2012.

\bibitem{7557075}
C.~Forster, L.~Carlone, F.~Dellaert, and D.~Scaramuzza, ``On-manifold
  preintegration for real-time visual--inertial odometry,'' \emph{IEEE
  Transactions on Robotics}, vol.~33, no.~1, pp. 1--21, Feb 2017.

\bibitem{munthe1999high}
H.~Munthe-Kaas, ``High order runge-kutta methods on manifolds,'' \emph{Applied
  Numerical Mathematics}, vol.~29, no.~1, pp. 115--127, 1999.

\bibitem{andrle2013geometric}
M.~S. Andrle and J.~L. Crassidis, ``Geometric integration of quaternions,''
  \emph{Journal of Guidance, Control, and Dynamics}, vol.~36, no.~6, pp.
  1762--1767, 2013.

\bibitem{li2005switched}
Z.~Li, Y.~Soh, and C.~Wen, \emph{Switched and impulsive systems: Analysis,
  design and applications}.\hskip 1em plus 0.5em minus 0.4em\relax Springer
  Science \& Business Media, 2005, vol. 313.

\bibitem{Hamel2002visual}
H.~Tarek and M.~Robert, ``Visual servoing of an under-actuated dynamic
  rigid-body system: An image-based approach,'' \emph{IEEE Transactions on
  Robotics and Automation}, vol.~18, no.~2, pp. 187--198, 2002.

\bibitem{mur2017visual}
R.~Mur-Artal and J.~D. Tard{\'o}s, ``Visual-inertial monocular slam with map
  reuse,'' \emph{IEEE Robotics and Automation Letters}, vol.~2, no.~2, pp.
  796--803, 2017.

\bibitem{horn2012matrix}
R.~A. Horn and C.~R. Johnson, \emph{Matrix analysis}.\hskip 1em plus 0.5em
  minus 0.4em\relax Cambridge university press, 2012.

\bibitem{kailath1980linear}
T.~Kailath, \emph{Linear systems}.\hskip 1em plus 0.5em minus 0.4em\relax
  Prentice-Hall Englewood Cliffs, NJ, 1980, vol. 156.

\bibitem{qin2017vins}
T.~Qin, P.~Li, and S.~Shen, ``Vins-mono: A robust and versatile monocular
  visual-inertial state estimator,'' \emph{IEEE Transactions on Robotics},
  vol.~34, no.~4, pp. 1004--1020, 2018.

\bibitem{zhang2018tutorial}
Z.~Zhang and D.~Scaramuzza, ``A tutorial on quantitative trajectory evaluation
  for visual (-inertial) odometry,'' in \emph{2018 IEEE/RSJ International
  Conference on Intelligent Robots and Systems (IROS)}.\hskip 1em plus 0.5em
  minus 0.4em\relax IEEE, 2018, pp. 7244--7251.

\bibitem{Burri25012016}
\BIBentryALTinterwordspacing
M.~Burri, J.~Nikolic, P.~Gohl, T.~Schneider, J.~Rehder, S.~Omari, M.~W.
  Achtelik, and R.~Siegwart, ``The euroc micro aerial vehicle datasets,''
  \emph{The International Journal of Robotics Research}, 2016. [Online].
  Available:
  \url{http://ijr.sagepub.com/content/early/2016/01/21/0278364915620033.abstract}
\BIBentrySTDinterwordspacing

\bibitem{zhang2014loam}
J.~Zhang and S.~Singh, ``Loam: Lidar odometry and mapping in real-time.'' in
  \emph{Robotics: Science and Systems}, vol.~2, 2014, p.~9.

\bibitem{mahmoud2014linear}
O.~E. Mahmoud, M.~R. Roman, and J.~F. Nasry, ``Linear and nonlinear stabilizing
  control of quadrotor uav,'' in \emph{2014 International Conference on
  Engineering and Technology (ICET)}.\hskip 1em plus 0.5em minus 0.4em\relax
  IEEE, 2014, pp. 1--8.

\bibitem{kocer2019model}
B.~B. Kocer, T.~Tjahjowidodo, and G.~G.~L. Seet, ``Model predictive uav-tool
  interaction control enhanced by external forces,'' \emph{Mechatronics},
  vol.~58, pp. 47--57, 2019.

\end{thebibliography}


\begin{thebibliography}{1}
\providecommand{\url}[1]{#1}
\csname url@samestyle\endcsname
\providecommand{\newblock}{\relax}
\providecommand{\bibinfo}[2]{#2}
\providecommand{\BIBentrySTDinterwordspacing}{\spaceskip=0pt\relax}
\providecommand{\BIBentryALTinterwordstretchfactor}{4}
\providecommand{\BIBentryALTinterwordspacing}{\spaceskip=\fontdimen2\font plus
\BIBentryALTinterwordstretchfactor\fontdimen3\font minus
  \fontdimen4\font\relax}
\providecommand{\BIBforeignlanguage}[2]{{%
\expandafter\ifx\csname l@#1\endcsname\relax
\typeout{** WARNING: IEEEtran.bst: No hyphenation pattern has been}%
\typeout{** loaded for the language `#1'. Using the pattern for}%
\typeout{** the default language instead.}%
\else
\language=\csname l@#1\endcsname
\fi
#2}}
\providecommand{\BIBdecl}{\relax}
\BIBdecl

\bibitem{7557075}
C.~Forster, L.~Carlone, F.~Dellaert, and D.~Scaramuzza, ``On-manifold
  preintegration for real-time visual--inertial odometry,'' \emph{IEEE
  Transactions on Robotics}, vol.~33, no.~1, pp. 1--21, Feb 2017.

\bibitem{horn2012matrix}
R.~A. Horn and C.~R. Johnson, \emph{Matrix analysis}.\hskip 1em plus 0.5em
  minus 0.4em\relax Cambridge university press, 2012.

\bibitem{kailath1980linear}
T.~Kailath, \emph{Linear systems}.\hskip 1em plus 0.5em minus 0.4em\relax
  Prentice-Hall Englewood Cliffs, NJ, 1980, vol. 156.

\bibitem{2li2015}
Z.~Li, K.~W. Wan, J.~Zheng, Z.~Zhu, Q.~Ren, and Y.~W. Yun, ``Real time
  localization of uav via fusion of visual signal and imu data,'' TD2015101,
  Tech. Rep., 2015.

\bibitem{john2019}
J.~Henawy, Z.~Li, W.~Yau, G.~Seet, and K.~Wan, ``Accurate imu preintegration
  using switched linear systems for autonomous systems,'' in \emph{2019 22nd
  International Conference on Intelligent Transportation Systems (ITSC)}.\hskip
  1em plus 0.5em minus 0.4em\relax IEEE, 2019, pp. 1--6.

\bibitem{li2005switched}
Z.~Li, Y.~Soh, and C.~Wen, \emph{Switched and impulsive systems: Analysis,
  design and applications}.\hskip 1em plus 0.5em minus 0.4em\relax Springer
  Science \& Business Media, 2005, vol. 313.

\end{thebibliography}
\bibliographystyle{IEEEtran}

\end{document}